\documentclass[letterpaper, 10 pt, conference]{ieeeconf}  

\IEEEoverridecommandlockouts                              

\usepackage{amsmath}
\usepackage{amssymb}
\usepackage{graphicx}
\usepackage{float}
\usepackage{array}
\usepackage{tikz}
\usepackage{listings}
\usepackage{mathrsfs}
\usepackage{pgfplots}
\usepackage{python}
\usepackage[pdfencoding=auto,psdextra,pagebackref,breaklinks,colorlinks]{hyperref}
\usepackage{graphicx}
\usepackage{cite}
\usepackage{dsfont}
\usepackage{mathtools,etoolbox}
\usepackage[none]{hyphenat}
\usepackage[utf8]{inputenc}
\usepackage[english]{babel}

\usepackage{amsthm}
\usepackage{xfrac}
\usepackage{nicefrac}
\usepackage{xcolor}
\usepackage{booktabs}
\usepackage[switch, pagewise]{lineno}
\usepackage{nicefrac}
\usepackage{flushend}
\usepackage{gensymb}
\usepackage{lipsum}
\usepackage{pgfplots}
\pgfplotsset{compat=1.18}

\DeclareUnicodeCharacter{2212}{-}
\usepackage{caption} 

\title{\LARGE \bf
\textit{MARVIS}: Motion \& Geometry Aware Real and Virtual Image Segmentation}

\author{Jiayi Wu*$^{1}$, Xiaomin Lin*$^{1}$, Shahriar Negahdaripour$^{1,2}$, Cornelia Fermüller$^{1}$, Yiannis Aloimonos$^{1}$ %
\thanks{This work was supported by USDA NIFA sustainable agriculture system program under award number 20206801231805.}%
\thanks{$^{1}$Maryland Robotics Center (MRC), University of Maryland, College Park, MD 20742, USA. 
Emails: \texttt{\{jiayiwu, xlin01, nshahriar, fermulcm, jyaloimo\}@umd.edu}.
}
\thanks{$^{2}$University of Miami, Coral Gables, FL 33146, USA, on sabbatical during the preparation of this paper.
Email: \texttt{nshahriar@miami.edu}.
}
\thanks{$^{*}$ represents equal contribution}
}

\begin{document}
\maketitle
\thispagestyle{empty}
\pagestyle{empty}

\begin{abstract}
Tasks such as autonomous navigation, 3D reconstruction, and object recognition near the water surfaces are crucial in marine robotics applications. However, challenges arise due to dynamic disturbances, e.g., light reflections and refraction from the random air-water interface, irregular liquid flow, and similar factors, which can lead to potential failures in perception and navigation systems. Traditional computer vision algorithms struggle to differentiate between real and virtual image regions, significantly complicating tasks. A virtual image region is an apparent representation formed by the redirection of light rays, typically through reflection or refraction, creating the illusion of an object's presence without its actual physical location. This work proposes a novel approach for segmentation on real and virtual image regions, exploiting synthetic images combined with domain-invariant information, a Motion Entropy Kernel, and Epipolar Geometric Consistency. Our segmentation network does not need to be re-trained if the domain changes. We show this by deploying the same segmentation network in two different domains: simulation and the real world. By creating realistic synthetic images that mimic the complexities of the water surface, we provide fine-grained training data for our network (MARVIS) to discern between real and virtual images effectively. By motion \& geometry-aware design choices and through comprehensive experimental analysis, we achieve state-of-the-art real-virtual image segmentation performance in unseen real world domain, achieving an IoU over $78\%$ and a $F_{1}$-Score over $86\%$ while ensuring a small computational footprint. MARVIS offers over $43$ FPS ($8$ FPS) inference rates on a single GPU (CPU core). Our code and dataset are available here \url{https://github.com/jiayi-wu-umd/MARVIS}. 
\end{abstract}
\section{INTRODUCTION}
\label{section:introduction}
\begin{figure}[t!]
\includegraphics[width=0.9\linewidth]{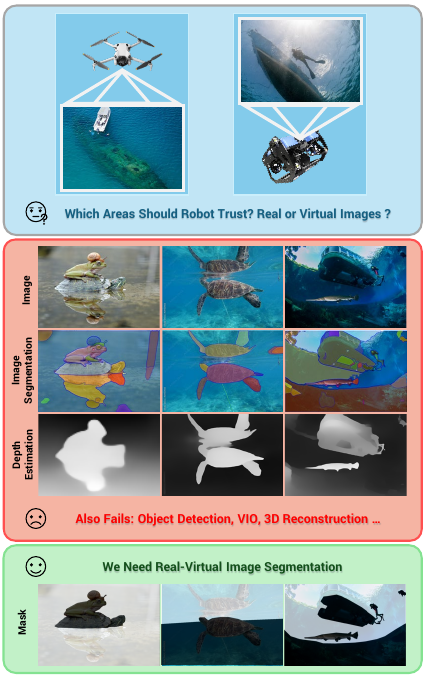}
\centering
\caption{Various computer vision tasks fail to varying degrees in multi-media scenarios due to virtual images formed by the diverging reflection or refraction of light rays. The real-virtual image segmentation mask is crucial to provide the robot with credible region information in the visual sensor input.}
\label{fig:Intro}
\vspace{-7mm}
\end{figure}

Advanced computer vision systems enable robots to comprehend scenes by analyzing pictorial information. Although this may work well in many cases, various real world scenarios can pose challenges due to the diverse nature of light propagation and scene complexity. The images received by human and machine visual sensors are often a complex superposition of direct imaging, reflections, refractions, scattering, and the likes. This complexity is particularly pronounced at the interfaces between different media. Consequently, the input to vision sensors contains real image regions formed by direct imaging and virtual areas typically formed by distortion, reflection, or refraction. Inherent complexities in differentiating between real and virtual regions pose challenges for both low-level and high-level computer vision tasks (see Fig. \ref{fig:Intro}), such as depth and pose estimations, visual odometry, object recognition, semantic segmentation, etc. Therefore, real-virtual image segmentation to localize visually credible areas is crucial for marine robots, particularly in multimedia scenarios. 

Novel deep learning-based methods have significantly improved the performance of numerous computer vision tasks such as object recognition \cite{he2016deep} and detection \cite{ren2016faster}, depth estimation \cite{10161471} and 3D navigation \cite{Zhou_2017_CVPR,wu2023low,10195131}.
However, our investigation indicates that real-virtual image segmentation in multimedia scenarios remains a relatively underexplored area. One primary challenge is the scarcity of publicly available datasets for real-virtual image segmentation in multimedia scenarios. Additionally, existing underwater simulators often overlook the intricate modeling of water-air interface imaging, making synthetic data extremely difficult to acquire. At the same time, data-driven methods are heavily relying on large-scale datasets. Secondly, the absence of an efficient annotation pipeline for the real-virtual image segmentation task, particularly for images with visual scene ambiguity, results in manual annotation being more laborious than common computer vision tasks. Moreover, achieving this task usually involves the use of Unmanned Aerial Vehicles (UAVs), Autonomous Surface Vehicles (ASVs), Unmanned Underwater Vehicles (UUVs), or even amphibious robots, which makes the acquisition of images from multiple domains costly.

To address these challenges, we develop \textit{AquaSim} to efficiently generate training data and simulate realistic underwater and water-air interface scenarios for related research. This comprehensive platform is tailored to train neural networks for water surface segmentation. Another key contribution is \textit{MARVIS}, a robust and efficient pipeline for real-virtual image segmentation.

Our proposed network incorporates temporal information and the epipolar geometric constraint into its supervised learning pipeline. We innovatively propose the Local Motion Entropy (LME) kernel, which serves as a pivotal layer in the encoder of the \textit{MARVIS} model. Based on the violation of the epipolar geometric constraint in the refractive virtual image area, we design an Epipolar Geometric Consistency (EGC) loss as weak supervision to further improve the performance of the pipeline when stereo image pairs are available. LME and EGC provide domain-invariant information, enabling robust segmentation in two domains (synthetic and real) without re-training.
With comprehensive motion \& geometry-aware learning on $3012$ pairs of synthetic images and their corresponding masks, \textit{MARVIS} achieves SOTA performance in both synthetic domain (IoU over $94\%$ and a F1-Score over $96\%$) and real world domain (IoU over $78\%$ and a F1-Score over $86\%$) despite having such a light architecture. Furthermore, \textit{MARVIS} offers significantly faster inference rates: over $43.4$ FPS on a NVIDIA™ RTX 4070 GPU and $8.0$ FPS on an Intel™ Core i9-4.10GHz CPU. 

This paper is structured as follows: Section~\ref{Section:related_work} contextualizes this work within previous research. Section~\ref{section:Aqua_Sim} outlines the simulator and the process for generating realistic images. Section~\ref{Section:model} explains in detail the structure of \textit{MARVIS}. Section~\ref{Section:Experiments_and_results} comprehensively evaluates our approach. Finally, Section~\ref{section:Conclusion} concludes with insights into future directions.
\section{BACKGROUND \& RELATED WORK}
\label{Section:related_work}
Autonomous robots play an increasingly vital role in environmental monitoring within the maritime sector. In addition to traditional challenges, the marine domain introduces unique obstacles. These include image distortions caused by water, artifacts, and virtual regions introduced by water turbulence and refraction, and the difficulty of collecting data to develop information-driven recognition and planning algorithms. To overcome these challenges, this paper introduces a simulator and a network that focus on robotic operations near the water surface.
\subsection{Underwater Simulators}

\begin{figure}[t!]
\vspace{3mm}
\includegraphics[width=0.85\linewidth]{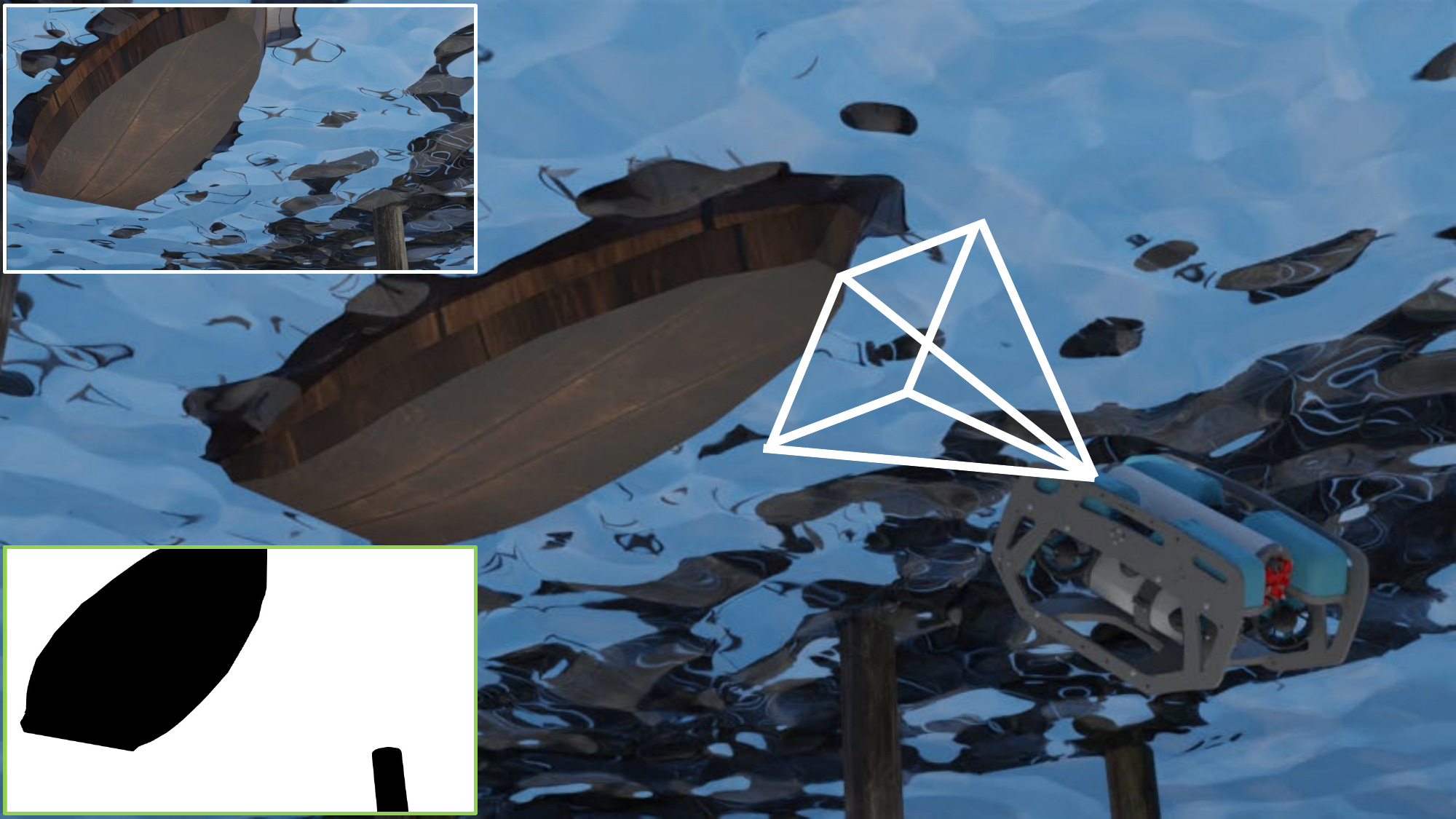}
\centering
\caption{BlueROV collecting data for our synthetic dataset in \textit{AquaSim}. The inset shows a sample image captured from the camera on a BlueROV. The top left displays the feed from a front-facing RGB
 camera, while the bottom left shows the real image's corresponding ground truth masks.}
\label{fig:BlueROV_in_sim}
\vspace{-5mm}
\end{figure}

Simulation environments are crucial for data collection in developing robotic capabilities. For example, advanced robotics simulators, such as IsaacSim\cite{liang2018gpu} and AirSim \cite{shah2018airsim}, offer high-fidelity rendering, particularly for aerial vehicles. Recent developments have addressed these critical needs for complex underwater environments and other maritime scenarios. For example, Zwilgmeyer et al. \cite{zwilgmeyer2021creating} use Blender to create underwater datasets, while UUV Simulator\cite{manhaes2016uuv} and UWSim \cite{dhurandher2008uwsim}provide platforms for modeling underwater physics and sensors. Despite this progress, these efforts have been discontinued. DAVE \cite{zhang2022dave} aims to fill this gap, but faces rendering limitations.

HoloOcean \cite{potokar2022holoocean}, MARUS \cite{lonvcar2022marus}, and UNav-Sim \cite{amer2023unav} have provided improved rendering realism but lack support to integrate essential robotics tools. Other simulators built on Unreal Engine often lack modifiability and have not yet released their original project files. ChatSim \cite{palnitkar2023chatsim} stands out for its integration of ChatGPT with OysterSim\cite{lin2022oystersim}, which offers intuitive modification of the simulated environment and the generation of photorealistic underwater environments. Notably, while the aforementioned simulators primarily target deep underwater tasks, they often lack flexibility in addressing real and virtual image segmentation across water surfaces. 

\subsection{Maritime and Marine Image Segmentation}
Several efforts \cite{chen2021wodis, bovcon2021wasr}
investigate sea-sky line inference and water obstacles detection for unmanned surface vehicles (USVs) using the Marine Semantic Segmentation Training Dataset (MaSTr1325), but do not explore the segmentation of virtual and real image regions,
Vandael et al. \cite{vandaele2021deep}, Muhadi et al. \cite{muhadi2021deep} and Wagner et al. \cite{wagner2023river}  worked on automated river level monitoring. However, these methods do not utilize information across frames and may not generalize effectively for tasks near the water surface, especially when observing from beneath the water surface. We propose a method that can work above and beneath the water surface and utilizes temporal cues in consecutive frames.  Unlike water segmentation at the water-air interface, real-virtual image segmentation prioritizes temporal motion and geometric details due to the robots' first-person viewpoint, egomotion, and proximity to water surfaces.

WaterNet by Liang et al. \cite{liang2020waternet} is the most similar work to our contribution, using the temporal cues in consecutive frames to track and segment volatile water. However, WaterNet \cite{liang2020waternet} relies on precise initial segmentation and lacks resilience to camera and object motion, leading to poor performance. Our method does not require prior knowledge and robust to camera and object motion.
Among other advancements in maritime and marine object segmentation, Sadrfaridpour et al. \cite{sadrfaridpour2021detecting} employ an underwater dataset for oyster detection. This relatively small dataset, which highlights the expensive and labor-intensive nature of data collection in underwater environments, prohibits generalization to natural oyster reefs. 
To address this challenge, we draw inspiration from various domains that use simulators to generate synthetic data, including oyster detection \cite{lin2023oysternet}, Remotely Operated Vehicle (ROV) detection \cite{lin2023seadronesim}, and whale detection \cite{gaur2023whale}. We created an underwater simulator that tailored a dataset suitable for real-virtual image segmentation.
\section{Underwater Simulator including Dynamic Water-air Interface Distortion}
\label{section:Aqua_Sim}

\begin{figure*}[t!]
\vspace{3mm}
\includegraphics[width=0.95\textwidth]{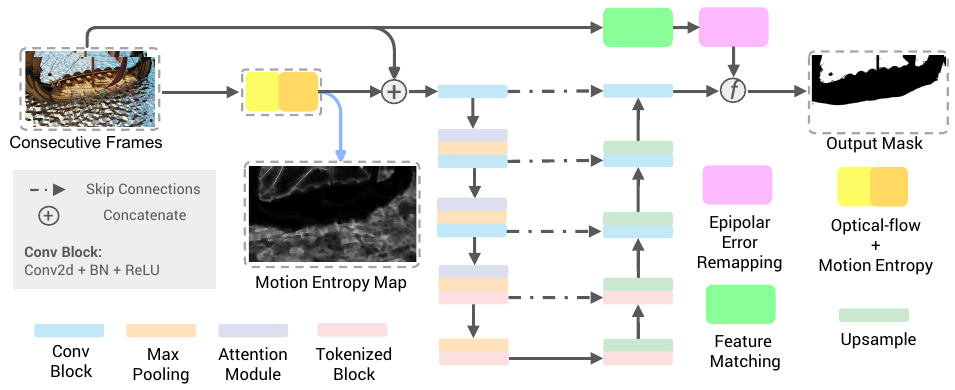}
\centering
\caption{Full learning pipeline of \textit{MARVIS}: initially, two consecutive grayscale video frames (at time $t-\delta t$ and $t$) are fed into optical flow estimation module \cite{teed2020raft}.  This motion information is forwarded to the proposed Local Motion Entropy (LME) layer for LME feature extraction. The fused spatio-temporal feature maps are then passed to a five-stage encoder to obtain features in high-dimensional latent space. Subsequently, the five-stage decoder progressively upsamples by fusing high-resolution feature maps from the corresponding stage encoder through skip connections to generate the segmentation mask with the original resolution. The learning objective involving pixel-wise losses (EGC loss) is given by Eq. \ref{eq:2_Epipolar_Loss}}
\label{fig:MARVIS_arch}
\vspace{-2mm}
\end{figure*}

\textit{AquaSim} leverages the Blender$^{\text{TM}}$ game engine to simulate environments for generating synthetic datasets (see Fig. \ref{fig:BlueROV_in_sim}) used in training neural networks for real-virtual image segmentation. By inputting desired parameters, such as water color, wave characteristics, and reflection properties (examples in Fig. 2, \textit{AquaSim}), users can create tailored datasets alongside corresponding ground-truth image masks. This streamlines the development of real-virtual image segmentation networks. 
Comparison with other simulators is detailed in Table \ref{Tab:comparision}. \textit{AquaSim} pioneers realistic water surfaces and uniquely provides ground-truth masks for real-virtual image alignment.
For maritime object detection, \textit{AquaSim} enables training neural networks to detect diverse shapes, colors, and oceanic variables like water texture, water color variations, different turbidity levels, and lighting conditions. Traditional datasets lack robustness due to limited underwater views, making \textit{AquaSim}'s provision of precise ground-truth image masks valuable for network training.

\subsection{\textit{AquaSim} Simulator}

\textbf{Water Modeling:} \textit{AquaSim} utilizes Blender$^{\text{TM}}$'s CYCLES rendering engine, employing path-tracing techniques for realistic oceanic environment rendering. Customizable features facilitate the creation of varied underwater scenes with different water properties, including color, texture, and turbidity levels. Additionally, \textit{AquaSim} offers real-time simulation capabilities via the EEVEE rendering engine, balancing rendering quality with interactivity.
\begin{table}[t]
\vspace{3mm}
\caption{Comparison of the different marine robotics simulators showing \textit{AquaSim} excels in rendering near-water surface}
\centering
\noindent\begin{tabular}{{l|}*{4}{c}}
\toprule
Simulator & Quality & Rendering  & Surface  & GT-Mask
\\
\midrule
UUV \cite{manhaes2016uuv}  & Low& Gazebo  & $\times$ & $\times$\\
URSim \cite{katara2019open} & Moderate & Unity3D & $\times$ & $\times$\\
UWRS \cite{chaudhary_2021}  & Moderate & Unity3D  & $\times$ & $\times$\\
HoloOcean \cite{potokar2022holoocean}  & Moderate & UE4  & $\times$& $\times$  \\
DAVE \cite{zhang2022dave} &Low &  Gazebo   & $\times$   & $\times$\\
MARUS \cite{lonvcar2022marus}  & Moderate & Unity3D  &\checkmark  & $\times$\\
UNav-Sim \cite{amer2023unav}  & High & UE5 & $\times$ & $\times$\\
\midrule
\textbf{\textit{AquaSim} (Ours)} & Highest & Blender &\checkmark &\checkmark \\
\bottomrule
\end{tabular}
\\\tiny{$^*$Only Blender and UE5 rendering engine supports ray-tracing with OptiX denoiser for photorealistic images; \\ Quality: Rendering quality;Rendering: Rendering Engine used; Surface: Realistic Water Surface Simulation; \\ GT-Mask: Ground truth real-virtual mask .}
\label{Tab:comparision}
\end{table}
A significant advantage of \textit{AquaSim} is its automated data generation that requires minimal user interaction. With a local installation of Blender, you can open an underwater scene equipped with pre-configured lighting and camera setups.

\textit{AquaSim} offers utilities for custom and prebuilt implementations, allowing users to fine-tune underwater environments. For example, users can adjust the water color using specific RGB value ranges, resulting in corresponding image renderings and masks. In addition, varying levels of visibility is achieved by turbidity adjustment.
Water texture is a key parameter in \textit{AquaSim}, offering customization options such as detail, dimension, scale, metallic properties, lacunarity, and strength, which influence the water surface appearance when viewed from below. Furthermore, \textit{AquaSim} incorporates white and Gaussian noise for the characteristics of the water. Users can also adjust underwater lighting conditions by specifying varying levels of sunlight penetration. Lastly, \textit{AquaSim} simulates various wave characteristics, such as height, tilt, sharpness, and textures, to further enhance the realism of the underwater scene.

\textbf{Sensor Modeling:}
In addition to generating synthetic underwater environments, \textit{AquaSim} offers the ability to generate data for the Inertial Measurement Unit (IMU) and information on the location of objects. This allows users to simulate object motions and accurately capture their trajectories using the following model:
\begin{equation}
    M_{sim} = M_{GT} + M_{bias} + M_{noise} 
\end{equation}

The simulated IMU measurement ($M_{\text{sim}}$) is composed of the ground truth IMU measurement ($M_{\text{GT}}$), the bias ($M_{\text{bias}}$), and the measurement noise ($M_{\text{noise}}$). The latter includes factors such as angle and velocity random walks and instability correlation bias. We have defined three error models that represent varying noise levels depending on the accuracy of the selected sensor. Lastly, we consider the vibration part of the noise, which can be characterized as white Gaussian, sinusoidal, or specified by its power spectral density.

Furthermore, \textit{AquaSim} automatically generates intrinsic calibration for both stereo cameras, ensuring an accurate representation of the simulated scene. This calibration process is essential for maintaining consistency between the virtual camera's parameters and real world camera characteristics, enhancing the realism of the generated images, and facilitating accurate data collection and analysis. 


\subsection{Dataset Generation}
Having described the \textit{AquaSim} capabilities, we next delve into the dataset creation process utilizing the enumerated objects within simulated scenes.
In this phase, \textit{AquaSim} systematically generates synthetic datasets comprising $60$ to $120$ image pairs for each scene. These feature a variety of objects: kayaks, canoes, rowboats, paddleboards, rafts, surfboards, jet skis, sailboats, fishing boats, beach balls, docks, buoys, water skiers, seaplanes, divers, lifebuoys, water bikes, windsurfers, floating platforms, floating restaurants, speedboats, submarines, yachts, pedal boats, anchors, lighthouses, crab traps, oyster beds, coral reefs, seagulls, and the likes.
The diverse array of objects embedded within a scene and the generation of multiple images per scene demonstrate \textit{ AquaSim}’s versatility in producing diverse datasets that encapsulate a wide range of maritime environments. This rich dataset composition facilitates robust training and evaluation of object detection algorithms,  thus enhancing accuracy in classification and recognition of objects commonly encountered in Aquatic environments. Furthermore, \textit{AquaSim} can also contribute high-quality datasets to advance research and development in the detection of maritime objects and other similar domains.

\section{Real and Virtual Image area Segmentation Pipeline}
\label{Section:model}
Our primary goal is to design a paradigm with robust and efficient performance in the real world, particularly with respect to camera and object motion. \textit{MARVIS} aims to perform in real-time with limited computing resources on the most widely used unmanned autonomous platforms.
\subsection{Motion Entropy Kernel} 

Virtual image areas generated by reflection and refraction often have static spatial features that are difficult to distinguish from the real image area. Therefore, spatial information alone does not provide sufficient visual cues to eliminate ambiguity in complex scenes. Moreover, air-water interfaces with highly variable dynamics and turbulence lead to virtual images with chaotic motions. In contrast, real objects in scenes within the same medium as the camera tend to exhibit relatively smooth motions in a sequence of consecutive frames. This suggests introducing a Local Motion Entropy (LME) measure as a key effective temporal feature for real-virtual image segmentation.
This paper proposes and considers it an essential and efficient temporal feature for real-virtual image segmentation. We propose LME as the following function:
\begin{equation}\label{eq:1_Motion_Entropy}
\begin{aligned}
H(\mathbf{M},\mathbf{A}) = & -\alpha \cdot\sum_{m \in \mathcal{M}}p(m)\log_{2} p(m) \\
& - \beta \cdot\sum_{a \in \mathcal{A}}p(a)\log_{2} p(a),
\end{aligned}
\end{equation}

where $\mathbf{M}$ and $\mathbf{A}$ are, respectively, the patch of motion vectors' magnitudes and angles within the local receptive field, $m$ denotes the possible values of the normalized motion vector's magnitude, $\mathcal{M}\rightarrow[0,1]$, and $a$ represents the possible values of the normalized motion vector angle, $\mathcal{A}\rightarrow[0,1]$. The probabilities $p(m)$ and $p(a)$ are calculated from the normalized histogram counts of the variables $m$ and $a$ separately, $\alpha$ and $\beta$ are the weights of the magnitude and angle of the motion vector in the calculation of local motion entropy, respectively (we set them both to 0.5 here).

Furthermore, we formulate the calculation of local motion entropy as a fixed parameter layer in the encoder of our \textit{MARVIS} pipeline and design a motion entropy kernel with an adjustable receptive field according to Eq. \ref{eq:1_Motion_Entropy}. Using the sliding-window algorithm, the motion entropy map of the entire image is calculated for different receptive field sizes. LME captures the relative motion entropy between consecutive frames, ensuring consistency despite camera or object motion. This property allows \textit{MARVIS} to be used flexibly in mobile robots rather than just fixed cameras. In addition, due to the property that local motion entropy is independent of pixel information, it is highly robust and consistent in different domains (e.g., underwater and terrestrial domains or synthetic and real world domains). This allows it to provide powerful and stable guidance for network training and improve the model’s zero-shot inference in unseen domains, as demonstrated in Section \ref{Section:Experiments_and_results}.
\subsection{Weak Supervision of Epipolar Geometric Consistency}
The propagating light refracts at the interface of two media with density difference, e.g., air and water. As a result, the refracted rays form a virtual image that no longer satisfies the epipolar geometry under different camera viewpoints. Based on this fact, we impose the epipolar geometric consistency error of sparse keypoints as weak supervision to improve the prediction accuracy of the pipeline, when stereo image pairs are available.
In addition to the sparsity of keypoint matches, the number of keypoints in different images is generally inconsistent. To build a suitable epipolar error map that serves as weak supervision of the network, we remap the normalized epipolar error values of keypoints back to the coordinates of the target image and set the pixel values of other non-keypoint pixels to zero. This ensures that the network can read the error map efficiently and operate in parallel. 
Accordingly, we define the Epipolar Geometric Consistency (EGC) loss based on the violation of epipolar geometric constraint in the refractive virtual image area:
\begin{equation}\label{eq:2_Epipolar_Loss}
\mathcal{L}_{EGC} = \frac{\sum[-(\hat{\mathbf{y}}-1) \cdot \mathbf{E}_{EGC}]}{Count_{(-(\hat{\mathbf{y}}-1) \cdot \mathbf{E}_{EGC}\neq 0)}},
\end{equation}
where $\hat{\mathbf{y}}$ is the predicted binary mask, and $\mathbf{E}_{EGC}$ represents the normalized epipolar error map.
\subsection{Network Architecture: \textit{MARVIS} Model}

\begin{figure}[t!]
\vspace{3mm}
\includegraphics[width=0.65\linewidth]{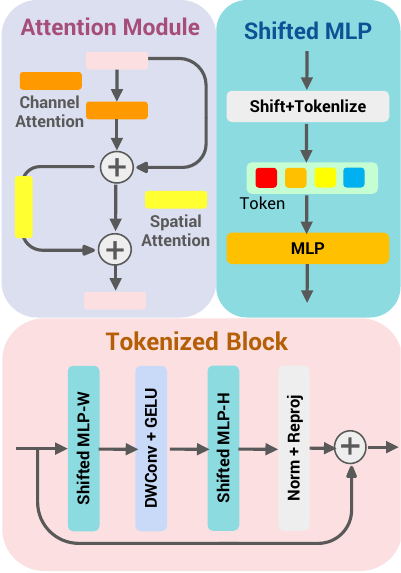}
\centering
\caption{The detailed operations in the Attention Module and the Tokenized Block (also described in Eq. \ref{eq:3_Tokenized_MLP}).}
\label{fig:Sub_module}
\end{figure}


The detailed network architecture is shown in Fig. \ref{fig:MARVIS_arch}. First, as input to the pipeline, two grayscale images from consecutive video frames (times  $t-\delta t$ and $t$) are passed to an optical flow estimation model, using RAFT here, to obtain motion information. The proposed motion entropy (ME) layer is applied to the optical flow to compute a local motion entropy map. The result and the input grayscale image at the current time t are concatenated and sent to the five-stage encoder of the \textit{MARVIS} model to extract features. The encoder contains three convolutional stages and two multilayer perceptron (MLP) stages. In the first convolution stage, the input that combines spatio-temporal information is passed through a convolution block with fewer filters for initial feature extraction. The resulting feature vector is fed into the two subsequent convolutional blocks with attention to further feature extraction. In the bottleneck, we use the tokenized MLP block \cite{valanarasu2022unext} to model a good representation in the latent space. The tokenized MLP blocks project the convolutional features into an abstract token with fewer dimensions and then utilize computationally lightweight MLPs to learn the feature and motion encodings for segmentation. The tokenized MLP block given in Fig. \ref{fig:Sub_module} performs the following operations:

\vspace{-3mm}
\begin{small} 
\begin{equation}\label{eq:3_Tokenized_MLP}
\begin{aligned}
\mathbf{Y} &= f(DWConv(MLP(Tokenize(Shift_{W}(\mathbf{X}))))) \\
\mathbf{Y} &= f(LN(\mathbf{T}+MLP(GELU(Tokenize(Shift_{H}(\mathbf{Y})))))),
\end{aligned}
\end{equation}
\end{small}
where $\mathbf{T}$ represents the tokens, $H$ and $W$ denote height and width, $DWConv$ is depth-wise convolution, $GELU$ \cite{hendrycks2023gaussian} is an activation function and LN is layer normalization.

Furthermore, lightweight Convolutional Block Attention Modules (CBAM) \cite{woo2018cbam} are integrated into stages $2$ to $4$ of the encoder, enhancing feature representation via channel and spatial attention. The \textit{MARVIS} model's decoder includes two tokenized MLP blocks and $3$ convolutional blocks. Each encoder stage downsamples feature resolution by $2$, while the decoder upsamples correspondingly. A skip connection links the encoder and decoder. Leveraging local motion entropy, \textit{MARVIS} efficiently learns latent representations, achieving strong segmentation performance with minimal parameters and computational complexity.
\begin{figure*}[t!]
\vspace{3mm}
\includegraphics[width=\textwidth]{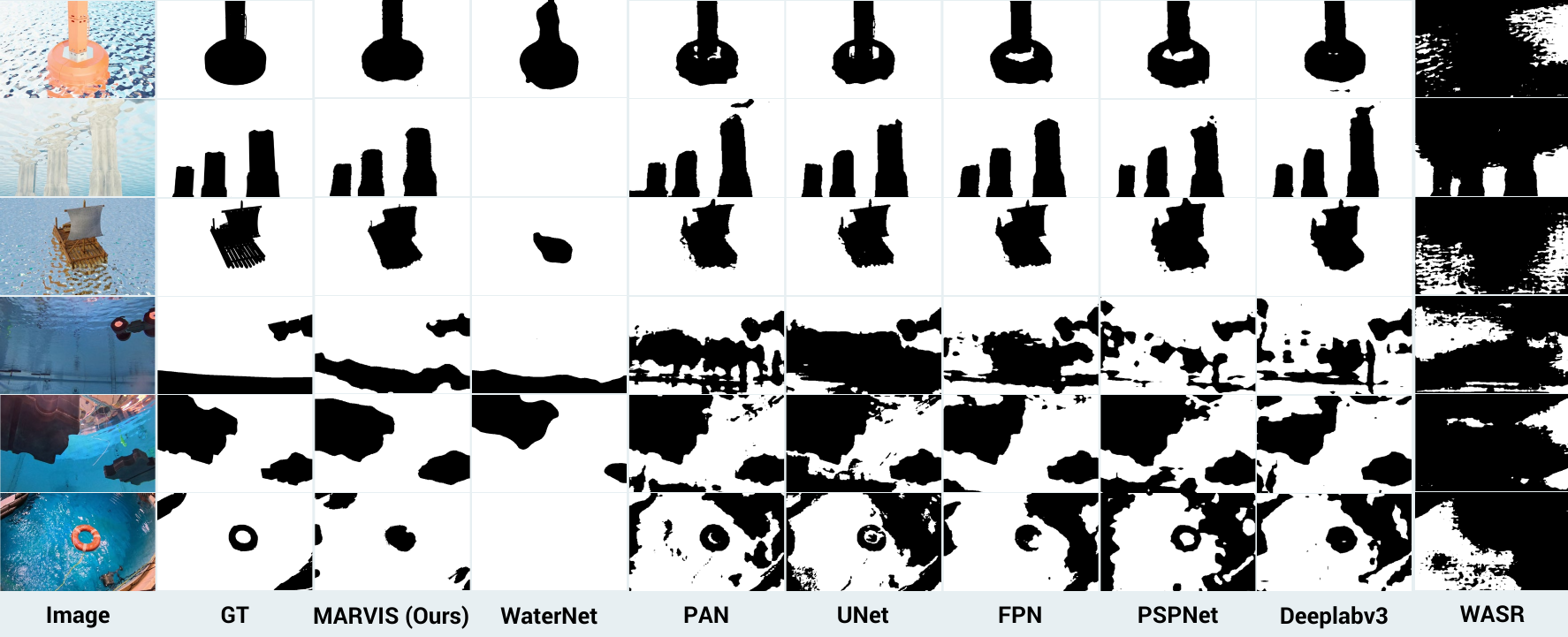}
\centering
\caption{A few qualitative comparisons are shown for real-virtual image segmentation by \textit{MARVIS} and SOTA water segmentation pipelines and widely used image segmentation models on both synthetic and real world testset. As seen, \textit{MARVIS} infers accurate and consistent segmentations across the samples from different domains.}
\vspace{-2mm}
\label{fig:Qual_results}
\end{figure*}
\subsection{Objective Function Formulation}
The end-to-end training is driven by the binary cross-entropy loss \cite{zhang2018generalized}, the Dice loss \cite{Sudre_2017} and the proposed EGC (epipolar geometric consistency) loss in Eq. \ref{eq:2_Epipolar_Loss} (when stereo image pairs are available). The cross-entropy loss quantifies the dissimilarity in pixel intensity distributions between the predicted segmentation mask ($\hat{y}$) and its ground truth ($y$). For a total of $n_{p}$ pixels, it is calculated as:
\begin{equation}\label{eq:4_BCE_Loss}
\mathcal{L}_{BCE} = \frac{1}{n_{p}}\sum_{i}[-y_{i}\log\hat{y}_{i}-(1-y_{i})\log(1-\hat{y}_{i})].
\end{equation}
To avoid class imbalance, we add the Dice loss, which balances real and virtual image classes by normalizing $y$ and $\hat{y}$ as follows:
\begin{equation}\label{eq:5_Dice_Loss}
\mathcal{L}_{Dice} = 1-\frac{2\sum_{i}y_{i}\hat{y}_{i}}{\sum_{i}y_{i}^{2}+\sum_{i}\hat{y}_{i}^{2}}
\end{equation}
Finally, the end-to-end learning objective is formulated as:
\begin{equation}\label{eq:6_Pipeline_Loss}
\mathcal{L}_{MARVIS} = \lambda_{B}\mathcal{L}_{BCE} + \lambda_{D}\mathcal{L}_{Dice} + \lambda_{E}\mathcal{L}_{EGC}.
\end{equation}
The optimal values $\lambda_{B} = 0.8$, $\lambda_{D} = 0.1$, and $\lambda_{E} = 0.1$.
in Eq. \ref{eq:6_Pipeline_Loss} have been determined empirically through hyper-parameter tuning.

\section{Experiments and Results}
\label{Section:Experiments_and_results}
Next, we describe the dataset and analyze segmentation experiments, where the models are trained on synthetic data and tested with real data and synthetic data.
\subsection{Dataset and Implementation Details}

\noindent
\textbf{Synthetic Data:}
Utilizing \textit{AquaSim} (described in Section \ref{section:Aqua_Sim}), we created a diverse synthetic dataset focusing on water turbidity, spacing of water waves, and air-water interface reflection. With varied lighting conditions, our dataset comprises 3012 pairs of $960 \times 540$ images with corresponding real-virtual masks.

\noindent
\textbf{Real Data: }
We conducted numerous missions at the National Robotics and Fisheries Research Institute (NBRF) at the University of Maryland (UMD). Two GoPro cameras on an ASV beneath the water surface were used to capture stereo images, and we labeled each image using the appropriate annotation tools. The dataset comprises 450 pairs of $960 \times 540$ images with corresponding real-virtual masks.  

\noindent
\textbf{Training Setups: }
\textit{MARVIS} is supervised on our synthetic dataset created with the AquaSim simulator. AquaSim automatically generates the ground truth of real-virtual image masks under different camera viewpoints. The training, validation, and test sets were split in a $80$:$5$:$15$ ratio and cross-validated during training. We used Pytorch libraries \cite{paszke2019pytorch} to implement the learning pipeline; The receptive field of the LME kernel is set to $7\times7$. AdamW \cite{loshchilov2019decoupled} is the optimizer with an initial learning rate of $0.001$ and exponential decay adjustment with a multiplicative factor of $0.9$. The threshold for the ratio test for feature matching is set to $0.7$. Our \textit{MARVIS} pipeline has about $2.56M$ parameters: $1.0M$ for the RAFT-S \cite{teed2020raft} optical flow estimator and $1.56M$ for the encoder and the decoder.

\subsection{Qualitative Evaluation}
To our knowledge, no other pipeline has been designed to perform the same real-virtual image segmentation task. For qualitative performance comparison, WaterNet \cite{liang2020waternet} and WASR \cite{bovcon2021wasr} are the most similar and may be considered as two SOTA water segmentation pipelines. As indicated in Fig. \ref{fig:Qual_results}, WASR's performance suffers as it addresses a different task. WaterNet typically excels initially, leveraging the initial ground-truth masks provided. However, it often struggles to sustain tracking of virtual image areas over time. Although WaterNet performs well for symmetric objects like the buoy (first row in Fig. \ref{fig:Qual_results}), it may lose track of non-symmetric moving objects such as the life buoy and toy car (fourth and sixth rows in Fig. \ref{fig:Qual_results}). All masks generated by WaterNet are inference images obtained after processing 50 or more consecutive frames.

Since our pipeline performs binary image segmentation, five other widely used SOTA image segmentation models are also relevant: Deeplabv3 \cite{chen2017rethinking}, PAN \cite{li2018pyramid}, UNet \cite{ronneberger2015unet}, FPN \cite{lin2017feature} and PSPNet \cite{zhao2017pyramid}. We trained these models by following their recommended settings on the same pipeline and train-validation data splits as \textit{MARVIS}. The qualitative performances of all these models for some samples are illustrated in Fig. \ref{fig:Qual_results}, leading to two main findings from the qualitative experimental analysis. 

First, training only on spatial pixel information is not effective for the specific task of real-virtual image segmentation, as we may conclude from Fig. \ref{fig:Qual_results}. Here, the trained network mispredicts many reflections and refractions as real images when composed of similar colors and textures that are inseparable in the RGB space. Moreover, WaterNet\cite{liang2020waternet} also employs motion information by frame differencing, which is no longer meaningful where both the camera and object(s) move. This leads to poor predictions in most samples captured by a moving camera. In contrast, \textit{MARVIS} provides higher-quality segmentation results by leveraging robust and consistent local motion entropy information in test scenarios and the violation of EGC by virtual images.

Second, all pipelines except \textit{MARVIS} give significantly worse segmentation results in real-world images than synthetic images. However, the performance of \textit{MARVIS} remains relatively consistent in both domains. This is attributed to training with only grayscale images and robust guidance from the LME layer for real-virtual image segmentation. This design reduces the domain gap between domains in the feature space, greatly improving the generalization to unseen domains.

\subsection{Quantitative Evaluation}
We use two standard metrics for quantitative assessments: Intersection over Union (IoU) \cite{minaee2020image} and $F_{1}$-Score \cite{10.1145/3606367}.  The IoU measures the performance of real-virtual image segmentation using the area of overlapping regions of the predicted and ground truth labels, defined as $IoU = \frac{Area\ of\ overlap}{Area\ of\ union}$. On the other hand, $F_{1}$-Score represents the balance of models that achieve high precision and high recall, which is defined as $F_{1} = \frac{2 \times Precision \times Recall}{Precision + Recall}$. The quantitative results are listed in Table \ref{table:quant_eval}.

The qualitative comparison and quantitative results are consistent and corroborate our analysis. For example, the real-world domain has scenes and lighting conditions that are different and more complex than those of the synthetic domain. According to Table \ref{table:quant_eval}, all pipelines degrade in performance by varying degrees in samples from the unseen real world. However, fully trained in the synthetic domain, the \textit{MARVIS} learning pipeline still achieves up to $78\%$ in IoU and $86\%$ in $F_{1}$-Score. This is a performance degradation of only $15.52\%$ (from $94\%$) in IoU and $9.88\%$ (from $96\%$) in $F_{1}$-Score, compared to $20.34\%-40.8\%$ in IoU and $13.75\%-28.11\%$ in $F_{1}$-Score by other pipelines

This confirms that the strategy to add the temporal information by the LME and EGC supervision (while ignoring color information) indeed enhances the zero-shot generalization ability of \textit{MARVIS} on the real-virtual image segmentation task. Furthermore, \textit{MARVIS} leverages the strong prior knowledge of LME and EGC designed for real-virtual image segmentation during the training process, enabling the lightweight network to learn a good feature representation for rapid convergence. \textit{MARVIS} has only $2.56M$ parameters that can rapidly generate segmentation masks at inference rates of $43.48$ FPS ($8.06$ FPS) on a single NVIDIA$^{\text{TM}}$ RTX 4070 GPU (Intel$^{\text{TM}}$ Core i9-4.10GHz CPU) core. Compared to other pipelines, it achieves much lower computational overhead and faster inference speeds, making it feasible to deploy on computationally constrained drones and AUVs.



\begin{table}[t!]
\vspace{3mm}
\centering
\begin{tabular}{c|c|c|c|c|c}
\toprule
Model & Params $\downarrow$ & \multicolumn{2}{c|}{IoU $\uparrow$} & \multicolumn{2}{c}{F1 $\uparrow$} \\ \cmidrule(lr){3-4} \cmidrule(lr){5-6}
 &  & Real & Syn & Real & Syn \\ \midrule
WASR \cite{bovcon2021wasr} & $71M$ & $13.24$ & $29.10$ & $21.78$ & $44.37$ \\
WaterNet \cite{liang2020waternet} & $22M$ & $41.08$ & $53.86$ & $49.63$ & $59.45$ \\
PSPNet \cite{zhao2017pyramid} & $11.32M$ & $61.35$ & $87.48$ & $74.63$ & $90.51$ \\
Deeplabv3 \cite{chen2017rethinking}& $5.63M$ & $68.68$ & $89.02$ & $79.26$ & $93.01$ \\
PAN \cite{li2018pyramid}& $4.10M$ & $61.65$ & $89.69$ & $74.45$ & $93.56$ \\
UNet \cite{ronneberger2015unet}& $31.04M$ & $51.11$ & $91.91$ & $66.85$ & $94.96$ \\
FPN \cite{lin2017feature}& $13.05M$ & $66.84$ & $91.91$ & $78.65$ & $94.94$ \\
\textbf{MARVIS(Ours)} & $\textbf{2.56M}$ & $\textbf{78.56}$ & $\textbf{94.08}$ & $\textbf{86.47}$ & $\textbf{96.35}$ \\ 
\bottomrule
\end{tabular}
\caption{Quantitative comparison for real-virtual image segmentation performance by \textit{MARVIS} and other SOTA water segmentation pipelines and widely used image segmentation models on both synthetic and real-world test sets.}
\label{table:quant_eval}
\end{table}

\section{Conclusion and Future Work}
\label{section:Conclusion}
We present a cutting-edge solution for real-virtual image segmentation near water surfaces, effectively leveraging synthetic data and domain-invariant features. By combining photorealistic 3D models with lightweight rendering techniques, \textit{MARVIS} enhances the diversity of the training dataset and improves segmentation results without the need for extensive manual labeling. Incorporating LME and EGC enhances the performance and robustness of the network in various environments. The successful demonstration of \textit{MARVIS'} effectiveness in both simulation and real-world scenarios underscores its potential to advance autonomous navigation, 3D reconstruction, and object recognition tasks in maritime applications. One possible improvement will be to perform real-virtual image segmentation in 3D. This information can be further used to perform 3D reconstruction in multi-media scenarios. 





\section*{ACKNOWLEDGMENTS}

We thank Michael Xu for helping with real data collection.


\bibliographystyle{IEEEtran}
\bibliography{IEEEabrv,refs}

\begin{thebibliography}{10}
\providecommand{\url}[1]{#1}
\csname url@rmstyle\endcsname
\providecommand{\newblock}{\relax}
\providecommand{\bibinfo}[2]{#2}
\providecommand\BIBentrySTDinterwordspacing{\spaceskip=0pt\relax}
\providecommand\BIBentryALTinterwordstretchfactor{4}
\providecommand\BIBentryALTinterwordspacing{\spaceskip=\fontdimen2\font plus
\BIBentryALTinterwordstretchfactor\fontdimen3\font minus \fontdimen4\font\relax}
\providecommand\BIBforeignlanguage[2]{{%
\expandafter\ifx\csname l@#1\endcsname\relax
\typeout{** WARNING: IEEEtran.bst: No hyphenation pattern has been}%
\typeout{** loaded for the language `#1'. Using the pattern for}%
\typeout{** the default language instead.}%
\else
\language=\csname l@#1\endcsname
\fi
#2}}

\bibitem{he2016deep}
K.~He, X.~Zhang, S.~Ren, and J.~Sun, ``Deep residual learning for image recognition,'' in \emph{Proceedings of the IEEE conference on computer vision and pattern recognition}, 2016, pp. 770--778.

\bibitem{ren2016faster}
S.~Ren, K.~He, R.~Girshick, and J.~Sun, ``Faster r-cnn: Towards real-time object detection with region proposal networks,'' \emph{IEEE Transactions on Pattern Analysis and Machine Intelligence}, vol.~39, no.~6, pp. 1137--1149, 2016.

\bibitem{10161471}
B.~Yu, J.~Wu, and M.~J. Islam, ``Udepth: Fast monocular depth estimation for visually-guided underwater robots,'' in \emph{2023 IEEE International Conference on Robotics and Automation (ICRA)}, 2023, pp. 3116--3123.

\bibitem{Zhou_2017_CVPR}
T.~Zhou, M.~Brown, N.~Snavely, and D.~G. Lowe, ``Unsupervised learning of depth and ego-motion from video,'' in \emph{Proceedings of the IEEE Conference on Computer Vision and Pattern Recognition (CVPR)}, July 2017.

\bibitem{wu2023low}
J.~Wu, ``Low-cost depth estimation and 3d reconstruction in scattering medium,'' Ph.D. dissertation, University of Florida, 2023.

\bibitem{10195131}
J.~Wu, B.~Yu, and M.~J. Islam, ``3d reconstruction of underwater scenes using nonlinear domain projection,'' in \emph{2023 IEEE Conference on Artificial Intelligence (CAI)}, 2023, pp. 359--361.

\bibitem{liang2018gpu}
J.~Liang, V.~Makoviychuk, A.~Handa, N.~Chentanez, M.~Macklin, and D.~Fox, ``Gpu-accelerated robotic simulation for distributed reinforcement learning,'' in \emph{Conference on Robot Learning}.\hskip 1em plus 0.5em minus 0.4em\relax PMLR, 2018, pp. 270--282.

\bibitem{shah2018airsim}
S.~Shah, D.~Dey, C.~Lovett, and A.~Kapoor, ``Airsim: High-fidelity visual and physical simulation for autonomous vehicles,'' in \emph{Field and Service Robotics: Results of the 11th International Conference}.\hskip 1em plus 0.5em minus 0.4em\relax Springer, 2018, pp. 621--635.

\bibitem{zwilgmeyer2021creating}
P.~G.~O. Zwilgmeyer, ``Creating a synthetic underwater dataset for egomotion estimation and 3d reconstruction,'' Master's thesis, NTNU, 2021.

\bibitem{manhaes2016uuv}
M.~M.~M. Manh{\~a}es, S.~A. Scherer, M.~Voss, L.~R. Douat, and T.~Rauschenbach, ``Uuv simulator: A gazebo-based package for underwater intervention and multi-robot simulation,'' in \emph{OCEANS 2016 MTS/IEEE Monterey}.\hskip 1em plus 0.5em minus 0.4em\relax IEEE, 2016, pp. 1--8.

\bibitem{dhurandher2008uwsim}
S.~K. Dhurandher, S.~Misra, M.~S. Obaidat, and S.~Khairwal, ``Uwsim: A simulator for underwater sensor networks,'' \emph{Simulation}, vol.~84, no.~7, pp. 327--338, 2008.

\bibitem{zhang2022dave}
M.~M. Zhang, W.-S. Choi, J.~Herman, D.~Davis, C.~Vogt, M.~McCarrin, Y.~Vijay, D.~Dutia, W.~Lew, S.~Peters, \emph{et~al.}, ``Dave aquatic virtual environment: Toward a general underwater robotics simulator,'' in \emph{2022 IEEE/OES Autonomous Underwater Vehicles Symposium (AUV)}.\hskip 1em plus 0.5em minus 0.4em\relax IEEE, 2022, pp. 1--8.

\bibitem{potokar2022holoocean}
E.~Potokar, S.~Ashford, M.~Kaess, and J.~G. Mangelson, ``Holoocean: An underwater robotics simulator,'' in \emph{2022 International Conference on Robotics and Automation (ICRA)}.\hskip 1em plus 0.5em minus 0.4em\relax IEEE, 2022, pp. 3040--3046.

\bibitem{lonvcar2022marus}
I.~Lon{\v{c}}ar, J.~Obradovi{\'c}, N.~Kra{\v{s}}evac, L.~Mandi{\'c}, I.~Kvasi{\'c}, F.~Ferreira, V.~Slo{\v{s}}i{\'c}, {\DJ}.~Na{\dj}, and N.~Mi{\v{s}}kovi{\'c}, ``Marus-a marine robotics simulator,'' in \emph{OCEANS 2022, Hampton Roads}.\hskip 1em plus 0.5em minus 0.4em\relax IEEE, 2022, pp. 1--7.

\bibitem{amer2023unav}
A.~Amer, O.~{\'A}lvarez-Tu{\~n}{\'o}n, H.~{\.I}. U{\u{g}}urlu, J.~L.~F. Sejersen, Y.~Brodskiy, and E.~Kayacan, ``Unav-sim: A visually realistic underwater robotics simulator and synthetic data-generation framework,'' in \emph{2023 21st International Conference on Advanced Robotics (ICAR)}.\hskip 1em plus 0.5em minus 0.4em\relax IEEE, 2023, pp. 570--576.

\bibitem{palnitkar2023chatsim}
A.~Palnitkar, R.~Kapu, X.~Lin, C.~Liu, N.~Karapetyan, and Y.~Aloimonos, ``Chatsim: Underwater simulation with natural language prompting,'' in \emph{OCEANS 2023-MTS/IEEE US Gulf Coast}.\hskip 1em plus 0.5em minus 0.4em\relax IEEE, 2023, pp. 1--7.

\bibitem{lin2022oystersim}
X.~Lin, N.~Jha, M.~Joshi, N.~Karapetyan, Y.~Aloimonos, and M.~Yu, ``Oystersim: Underwater simulation for enhancing oyster reef monitoring,'' in \emph{OCEANS 2022, Hampton Roads}.\hskip 1em plus 0.5em minus 0.4em\relax IEEE, 2022, pp. 1--6.

\bibitem{chen2021wodis}
X.~Chen, Y.~Liu, and K.~Achuthan, ``Wodis: Water obstacle detection network based on image segmentation for autonomous surface vehicles in maritime environments,'' \emph{IEEE Transactions on Instrumentation and Measurement}, vol.~70, pp. 1--13, 2021.

\bibitem{bovcon2021wasr}
B.~Bovcon and M.~Kristan, ``Wasr—a water segmentation and refinement maritime obstacle detection network,'' \emph{IEEE Transactions on Cybernetics}, vol.~52, no.~12, pp. 12\,661--12\,674, 2021.

\bibitem{vandaele2021deep}
R.~Vandaele, S.~L. Dance, and V.~Ojha, ``Deep learning for automated river-level monitoring through river-camera images: an approach based on water segmentation and transfer learning,'' \emph{Hydrology and Earth System Sciences}, vol.~25, no.~8, pp. 4435--4453, 2021.

\bibitem{muhadi2021deep}
N.~A. Muhadi, A.~F. Abdullah, S.~K. Bejo, M.~R. Mahadi, and A.~Mijic, ``Deep learning semantic segmentation for water level estimation using surveillance camera,'' \emph{Applied Sciences}, vol.~11, no.~20, p. 9691, 2021.

\bibitem{wagner2023river}
F.~Wagner, A.~Eltner, and H.-G. Maas, ``River water segmentation in surveillance camera images: A comparative study of offline and online augmentation using 32 cnns,'' \emph{International Journal of Applied Earth Observation and Geoinformation}, vol. 119, p. 103305, 2023.

\bibitem{liang2020waternet}
Y.~Liang, N.~Jafari, X.~Luo, Q.~Chen, Y.~Cao, and X.~Li, ``Waternet: An adaptive matching pipeline for segmenting water with volatile appearance,'' \emph{Computational Visual Media}, vol.~6, pp. 65--78, 2020.

\bibitem{sadrfaridpour2021detecting}
B.~Sadrfaridpour, Y.~Aloimonos, M.~Yu, Y.~Tao, and D.~Webster, ``Detecting and counting oysters,'' in \emph{2021 IEEE International Conference on Robotics and Automation (ICRA)}.\hskip 1em plus 0.5em minus 0.4em\relax IEEE, 2021, pp. 2156--2162.

\bibitem{lin2023oysternet}
X.~Lin, N.~J. Sanket, N.~Karapetyan, and Y.~Aloimonos, ``Oysternet: Enhanced oyster detection using simulation,'' in \emph{2023 IEEE International Conference on Robotics and Automation (ICRA)}.\hskip 1em plus 0.5em minus 0.4em\relax IEEE, 2023, pp. 5170--5176.

\bibitem{lin2023seadronesim}
X.~Lin, C.~Liu, A.~Pattillo, M.~Yu, and Y.~Aloimonous, ``Seadronesim: Simulation of aerial images for detection of objects above water,'' in \emph{Proceedings of the IEEE/CVF Winter Conference on Applications of Computer Vision}, 2023, pp. 216--223.

\bibitem{gaur2023whale}
A.~Gaur, C.~Liu, X.~Lin, N.~Karapetyan, and Y.~Aloimonos, ``Whale detection enhancement through synthetic satellite images,'' in \emph{OCEANS 2023-MTS/IEEE US Gulf Coast}.\hskip 1em plus 0.5em minus 0.4em\relax IEEE, 2023, pp. 1--7.

\bibitem{teed2020raft}
Z.~Teed and J.~Deng, ``Raft: Recurrent all-pairs field transforms for optical flow,'' 2020.

\bibitem{katara2019open}
P.~Katara, M.~Khanna, H.~Nagar, and A.~Panaiyappan, ``Open source simulator for unmanned underwater vehicles using ros and unity3d,'' in \emph{2019 IEEE Underwater Technology (UT)}.\hskip 1em plus 0.5em minus 0.4em\relax IEEE, 2019, pp. 1--7.

\bibitem{chaudhary_2021}
A.~Chaudhary, R.~Mishra, B.~Kalyan, and M.~Chitre, ``Development of an underwater simulator using unity3d and robot operating system,'' in \emph{{OCEANS} 2021 {MTS}/{IEEE}}.\hskip 1em plus 0.5em minus 0.4em\relax {IEEE}, Sep 2021.

\bibitem{valanarasu2022unext}
J.~M.~J. Valanarasu and V.~M. Patel, ``Unext: Mlp-based rapid medical image segmentation network,'' 2022.

\bibitem{hendrycks2023gaussian}
D.~Hendrycks and K.~Gimpel, ``Gaussian error linear units (gelus),'' 2023.

\bibitem{woo2018cbam}
S.~Woo, J.~Park, J.-Y. Lee, and I.~S. Kweon, ``Cbam: Convolutional block attention module,'' 2018.

\bibitem{zhang2018generalized}
Z.~Zhang and M.~R. Sabuncu, ``Generalized cross entropy loss for training deep neural networks with noisy labels,'' 2018.

\bibitem{Sudre_2017}
C.~H. Sudre, W.~Li, T.~Vercauteren, S.~Ourselin, and M.~Jorge~Cardoso, \emph{Generalised Dice Overlap as a Deep Learning Loss Function for Highly Unbalanced Segmentations}.\hskip 1em plus 0.5em minus 0.4em\relax Springer International Publishing, 2017, p. 240–248.

\bibitem{paszke2019pytorch}
A.~Paszke, S.~Gross, F.~Massa, A.~Lerer, J.~Bradbury, G.~Chanan, T.~Killeen, Z.~Lin, N.~Gimelshein, L.~Antiga, A.~Desmaison, A.~Köpf, E.~Yang, Z.~DeVito, M.~Raison, A.~Tejani, S.~Chilamkurthy, B.~Steiner, L.~Fang, J.~Bai, and S.~Chintala, ``Pytorch: An imperative style, high-performance deep learning library,'' 2019.

\bibitem{loshchilov2019decoupled}
I.~Loshchilov and F.~Hutter, ``Decoupled weight decay regularization,'' 2019.

\bibitem{chen2017rethinking}
L.-C. Chen, G.~Papandreou, F.~Schroff, and H.~Adam, ``Rethinking atrous convolution for semantic image segmentation,'' 2017.

\bibitem{li2018pyramid}
H.~Li, P.~Xiong, J.~An, and L.~Wang, ``Pyramid attention network for semantic segmentation,'' 2018.

\bibitem{ronneberger2015unet}
O.~Ronneberger, P.~Fischer, and T.~Brox, ``U-net: Convolutional networks for biomedical image segmentation,'' 2015.

\bibitem{lin2017feature}
T.-Y. Lin, P.~Dollár, R.~Girshick, K.~He, B.~Hariharan, and S.~Belongie, ``Feature pyramid networks for object detection,'' 2017.

\bibitem{zhao2017pyramid}
H.~Zhao, J.~Shi, X.~Qi, X.~Wang, and J.~Jia, ``Pyramid scene parsing network,'' 2017.

\bibitem{minaee2020image}
S.~Minaee, Y.~Boykov, F.~Porikli, A.~Plaza, N.~Kehtarnavaz, and D.~Terzopoulos, ``Image segmentation using deep learning: A survey,'' 2020.

\bibitem{10.1145/3606367}
P.~Christen, D.~J. Hand, and N.~Kirielle, ``A review of the f-measure: Its history, properties, criticism, and alternatives,'' \emph{ACM Comput. Surv.}, vol.~56, no.~3, oct 2023.

\end{thebibliography}

\end{document}